\documentclass{ieeeaccess}
\usepackage{lipsum,multicol}
\usepackage[export]{adjustbox}
\usepackage{url}
\usepackage{textcomp}
\usepackage{amsmath}
\usepackage{pifont}
\usepackage{caption}
\usepackage{color, colortbl}
\usepackage{xcolor}
\usepackage{array}
\usepackage{dingbat}
\usepackage{multirow}
\usepackage[cmintegrals]{newtxmath}
\usepackage{cite}
\usepackage{graphics}
\usepackage{graphicx}
\usepackage{subfigure}
\usepackage{enumitem}
\newcommand{\rev}[1]{\textcolor{black}{#1}} 
\newcolumntype{C}[1]{>{\centering\let\newline\\\arraybackslash\hspace{0pt}}m{#1}}
\newcolumntype{L}[1]{>{\raggedright\let\newline\\\arraybackslash\hspace{0pt}}m{#1}}
\def\BibTeX{{\rm B\kern-.05em{\sc i\kern-.025em b}\kern-.08em
    T\kern-.1667em\lower.7ex\hbox{E}\kern-.125emX}}
    
\begin{document}

\doi{10.1109/ACCESS.2020.3032344}

\title{Driver Behavior Recognition via Interwoven Deep Convolutional Neural Nets with Multi-stream Inputs}
\author{\uppercase{Chaoyun Zhang}\authorrefmark{1},
\uppercase{Rui Li\authorrefmark{2}, Woojin Kim\authorrefmark{3}, Daesub Yoon\authorrefmark{3} and Paul~Patras\authorrefmark{1}},
\IEEEmembership{Senior Member, IEEE}}
\address[1]{Institute for Computing Systems Architecture (ICSA), School of Informatics, University of Edinburgh, UK. (e-mails: \{chaoyun.zhang, paul.patras\}@ed.ac.uk)}
\address[2]{Samsumg AI Cambridge Center, Cambridge, UK. (e-mail: rui.li@samsung.com)}
\address[3]{Electronics and Telecommunications Research Institute (ETRI),
Daejon, South Korea. (e-mails: \{wjinkim, eyetracker\}@etri.re.kr.)}

\begin{abstract}
Understanding driver activity is vital for in-vehicle systems that aim to reduce the incidence of car accidents rooted in cognitive distraction. Automating \textit{real-time} behavior recognition while ensuring actions classification with \textit{high accuracy} is however challenging, given the multitude of circumstances surrounding drivers, the unique traits of individuals, and the computational constraints imposed by in-vehicle embedded platforms. Prior work fails to jointly meet these runtime/accuracy requirements and mostly rely on a single sensing modality, which in turn can be a single point of failure.
In this paper, we harness the exceptional feature extraction abilities of deep learning and propose a dedicated Interwoven Deep Convolutional Neural Network (InterCNN) architecture to tackle the problem of accurate classification of driver behaviors in real-time. The proposed solution exploits information from multi-stream inputs, i.e., in-vehicle cameras with different fields of view and optical flows computed based on recorded images, and merges through multiple fusion layers abstract features that it extracts. This builds a tight ensembling system, which significantly improves the robustness of the model. In addition, we introduce a temporal voting scheme based on historical inference instances, in order to enhance the classification accuracy. Experiments conducted with a dataset that we collect in a mock-up car environment demonstrate that the proposed InterCNN with MobileNet convolutional blocks can classify 9 different behaviors with 73.97\% accuracy, and 5 `aggregated' behaviors with 81.66\% accuracy. We further show that our architecture is highly computationally efficient, as it performs inferences within 15~ms, which satisfies the real-time constraints of intelligent~cars. Nevertheless, our InterCNN is robust to lossy input, as the classification remains accurate when two input streams are occluded.
\end{abstract}

\begin{keywords}
Driver behavior recognition, deep learning, convolutional neural networks 
\end{keywords}
\titlepgskip=-15pt
\maketitle

\section{Introduction}
\IEEEPARstart{D}{river}'s cognitive distraction is a major cause of unsafe driving, which leads to severe car accidents every year~\cite{liao2016detection}. Actions that underlie careless driving include interacting with passengers, using a mobile phone (e.g., for text messaging, game playing, and web browsing), and consuming food or drinks \cite{stutts2001role}. Such behaviors contribute significantly to delays in driver's response to unexpected events, thereby increasing the risk of collisions. Identifying driver behaviors is therefore becoming increasingly important for car manufacturers, who aim to build in-car intelligence that can improve safety by notifying drivers in real-time of potential hazards \cite{kaplan2015driver}. Further, although full car automation is still years ahead, inferring driver behaviors is essential for vehicles with partial (``hands off'') and conditional (``eyes off'') automation, which will dominate the market at least until 2030~\cite{kpmg}. This is because the driver must either be ready to take control at any time or intervene in situations where the vehicle cannot complete certain critical functions~\cite{sae}.

\begin{figure}[t]
\begin{center}
\includegraphics[width=\columnwidth]{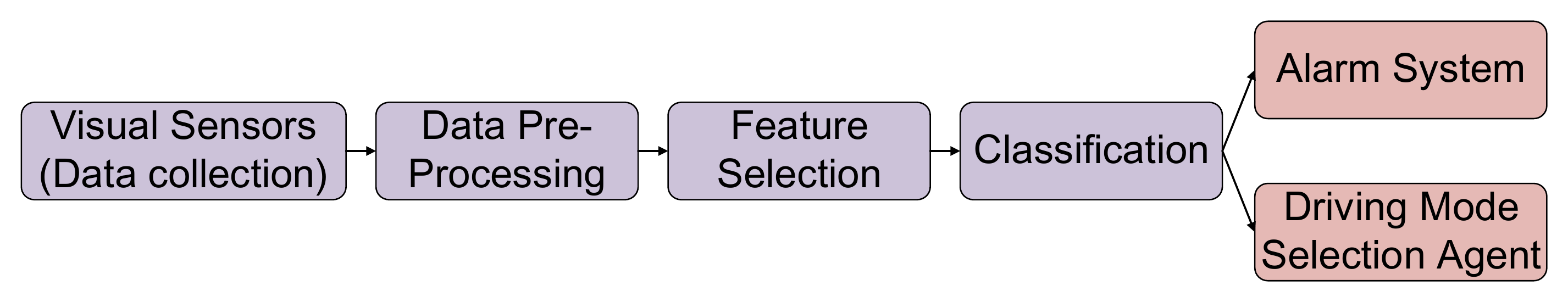}
\end{center}
\caption{\label{fig:system} The typical pipeline of driver behavior classification and alarm/driving mode selection systems.}
\end{figure}

Modern driver behavior classification systems usually rely on videos acquired from in-vehicle cameras, which record the movements and facial expressions of the driver~\cite{kutila2007driver}. The videos captured are routinely partitioned into sequences of image frames, which are then pre-processed for features selection~\cite{kulkarni2017review}. Such features are fed to pre-trained classifiers to perform identification of different actions that the driver performs. Subsequently, the classifier may trigger an alarm system in manual driving cars or provide input to a driving mode selection agent in semi-autonomous vehicles. We illustrate this pipeline in Figure~\ref{fig:system}. During this process, the accuracy of the classifier is key to the performance of the system. In addition, the system should perform such classification in real-time, so as to help the driver mitigate unsafe circumstances in a timely manner. Achieving high accuracy while maintaining runtime efficiency is however challenging. Striking appropriate trade-offs between these aims is therefore vital for intelligent and autonomous vehicles.

Underpinned by recent advances in parallel computing, deep neural networks \cite{lecun2015deep} have achieved remarkable results in various areas, including computer vision \cite{8677269}, control \cite{mnih2015human}, and autonomous driving \cite{8481390, 8481511}, as they can automatically extract features from raw data without requiring expensive hand-crafted feature engineering. Graphics processing units (GPUs) allow to train deep neural networks rapidly and with great accuracy, and perform inferences fast. Moreover, System on Chip (SoC) designs optimized for mobile artificial intelligence (AI) applications are becoming more powerful and computationally efficient~\cite{zhang2018deep}, and embedding deep learning in car systems increasingly affordable \cite{nvidia}. Therefore, potential exists to build high precision driver behavior classification systems without compromising runtime performance. However, recent works that adopt deep learning to solve the driver activity recognition problem, including \cite{alotaibi2020,xing2019,kose2019,leekha2019,majdi2018drive,tran2018real,valeriano2018recognition}, suffer from at least one of the following limitations: \textit{(i)} they do not quantify the inference times of the solution proposed, which is critical to real-life car systems, or exhibit runtimes that are not affordable in practice; \textit{(ii)} often struggle to classify individual actions with very high accuracy; and \textit{(iii)} rely on a single sensing modality (video feed) for detection, which can become a single point of failure, thus challenging the practical efficacy of the classifier.

To tackle these problems, in this paper we design a driver behavior recognition system that uniquely combines different convolutional-type neural models, through which we accurately perform this task in real-time, relying on multiple inputs. As such, we make the following \textbf{key contributions}:
\begin{enumerate}
    \item We build a mock-up environment to emulate self-driving car conditions and instrument a detailed user study for data collection purposes. Specifically, we deploy side and front facing cameras to record the body movements and facial expressions of 50 participant drivers, as they perform a range of tasks. This leads to a large driver behavior video dataset recorded from two angles (front and side), which we use for model training and evaluation. To the best of knowledge, this constitutes the first dual-view dataset for driver behavior classification.
    \item We architect original Interwoven Convolutional Neural Networks (InterCNNs) to perform feature extraction and fusions across multiple levels, based on multi-stream video inputs and optical flow information. Our novel design allows to extract in parallel and mix increasingly abstract features, as well as plugging in different lightweight CNN architectures (e.g. MobileNet \cite{howard2017mobilenets, sandler2018mobilenetv2}), enabling to determine which building blocks can improve the computation efficiency of in-vehicle systems.
    \item We demonstrate that our InterCNNs with MobileNet blocks and a temporal voting scheme, which enhances accuracy by leveraging historical inferences, can classify 9 different behaviors with 73.97\% accuracy, and 5 aggregated behaviors (i.e., grouping tasks that involve the use of a mobile device, and eating and drinking) with 81.66\% accuracy. Our architecture can make inferences within 15~ms, which satisfies the timing constraints posed by real car systems. Importantly, our architecture is highly robust to lossy input, as it remains accurate when two streams of the input are occluded.
\end{enumerate}

The results obtained demonstrate the feasibility of using deep learning to accurately infer driver behavior in real-time, thereby making important steps towards fulfilling the multi-trillion economic potential of the driverless car industry~\cite{fortune}.

The rest of the paper is organized as follows. In Sec.~\ref{sec:related}, we discuss relevant related work. In Sec.~\ref{sec:data}, we present our data collection and pre-processing efforts, which underpin the design of our neural network solution that we detail in Sec.~\ref{sec:model}. We demonstrate the performance of the proposed InterCNNs by means of experiments reported in Sec.~\ref{Sec:exp}. Sec.~\ref{sec:conclusions} concludes our contributions.

\section{Related Work}
\label{sec:related}
The majority of driver behavior classification systems are based on in-vehicle vision instruments (i.e., cameras or eye-tracking devices), which constantly monitor the movements of the driver \cite{fernandez2016driver}. The core of such systems is therefore tasked with a computer vision problem, whose objective is to classify actions performed by drivers, using sequences of images acquired in real-time. Existing research can be categorized into two main classes: non deep learning and deep learning approaches. 

\subsection{Non Deep Learning Based Driver Behavior Identification}
In \cite{liu2016driver}, Liu \emph{et al.} employ Laplacian Support Vector Machine (SVM) and extreme learning machine techniques to detect driver distraction, using labelled data that captures vehicle dynamic and drivers' eye and head movements. Experiments show that this semi-supervised approach can achieve up to 97.2\% detection accuracy. Li \emph{et al.} pursue distraction detection from a different angle. They exploit kinematic signals from the vehicle Controller Area Network (CAN) bus, to alleviate the dependency on expensive vision sensors. Detection is then performed with an SVM, achieving 95\% accuracy.

Ragab \emph{et al.} compare the prediction accuracy of different machine learning methods in driving distraction detection \cite{ragab2014visual}, showing that Random Forests perform best and require only 0.05~s per inference. Liao \emph{et al.} consider drivers' distraction in two different scenarios, i.e., stop-controlled intersections and speed-limited highways~\cite{liao2016detection}. They design an SVM classifier operating with Recursive Feature Elimination (RFE) to detect distraction while driving. The evaluation results suggest that by fusing eye movements and driving performance information, the classification accuracy can be improved in stop-controlled intersection settings.

\subsection{Deep Learning Based Driver Behavior Identification}
Deep learning is becoming increasingly popular for identifying driver behaviors. In \cite{hoang2016multiple}, a multiple scale Faster Region CNN is employed to detect whether a driver is using a mobile phone or their hands are on the steering wheel. The solution operates on images of the face, hands and steering wheel separately, and then performs classification on these regions of interest. Experimental results show that this model can discriminate behaviors with high accuracy in real-time. Majdi \emph{et al.} design a dedicated CNN architecture called Drive-Net to identify 10 different behaviors of distracted driving \cite{majdi2018drive}. Experiments suggest that applying Region of Interest (RoI) analysis on images of faces can significantly improve  accuracy.
A CNN model is also adopted recently in \cite{leekha2019}, where the authors show the addition of features including driver posture through foreground extraction can improve classification accuracy. Similarly, Kose \emph{et al.} reveal that augmenting the input to CNN structures with temporal information can further enhance classification performance~\cite{kose2019}. 

Tran \emph{et al.} build a driving simulator named Carnetsoft to collect driving data, and utilize 4 different CNN architectures to identify 10 distracted and non-distracted driving behaviors \cite{tran2018real}. The authors observe that deeper architectures can improve the detection accuracy, but at the cost of increased inference times. Investigating the trade-off between accuracy and efficiency remains an open issue. The performance of three different pre-trained convolutional structures, i.e., AlexNet, GoogLeNet, and ResNet50 is compared in~\cite{xing2019} on a similar binary classification task, revealing all candidates yield practical inference times in a Matlab-based environment. A comparison of different neural models applied to driver behavior classification is also carried out in~\cite{alotaibi2020}, where the authors propose an ensemble of convolutional and hierarchical recurrent neural networks, which increases classification accuracy at the cost of increased runtimes.

Yuen \emph{et al.} employ a CNN to perform head pose estimation during driving \cite{yuen2016looking}. Evaluation results suggest that incorporating a Stacked Hourglass Network to estimate landmarks and refine the face detection can significantly improve the accuracy with different occlusion levels. In \cite{streiffer2017darnet}, Streiffer \emph{et al.} investigate mixing different models, i.e., CNNs, recurrent neural networks (RNNs), and SVMs, to detect driver distraction. Their ensembling CNN + RNN approach significantly outperforms simple CNNs in terms of prediction accuracy.
Valeriano \emph{et al.} combine live video and optical flows to classify driver behaviors \cite{valeriano2018recognition}, using a two stream inflated 3D CNN (I3D) \cite{carreira2017quo}. This work trains two I3Ds for video flow and optical flow separately and averages their predictions at test time. 

Recognizing driver's behavior with high accuracy, using inexpensive sensing infrastructure, and achieving this in real-time remain challenging, yet mandatory for intelligent vehicles that can improve safety and reduce the time during which the driver is fully engaged. To the best of our knowledge, existing work fails to meet all these requirements and only performs behavior recognition from a single view. 

\section{Data Collection and Pre-Processing}\label{sec:data}
In this work, we propose an original driver behavior recognition system that can classify user actions accurately in real-time, using input from in-vehicle cameras. Before delving into our solution (Sec.~\ref{sec:model}), we discuss the data collection and pre-processing campaign that we conduct while mimicking an autonomous vehicle environment, in order to facilitate the design, training, and evaluation of our neural network model.

\subsection{Why a Dual-view Driver Behavior Classification Dataset}
Existing datasets previously employed for driver behavior classification (e.g. \cite{abouelnaga2017real, kaggle, Naqvi2018}) are largely recorded from a single viewpoint, i.e., only one camera records the driver's actions. Such camera is typically installed below the roof level on the passenger's side, in order to provide a comprehensive view of the driver's body movements. However, this camera alone cannot capture clearly the driver's facial expressions, which may reflect their cognitive status. Incorporating a second camera front-facing the driver is therefore important, as it gathers information from a different angle, providing extra features that can expedite learning and improving the performance of the classification system. Importantly,
in the event that one camera is occluded or malfunctioning, the driver behavior recognition instrument can continue to operate and provide the intended functionality, as our results will demonstrate. Aiming to achieve these goals prompts us to generate a fresh dataset in which the state of the drivers is recorded from different angles (i.e., capturing both body movements and facial expressions), so as to enable building a highly accurate classification framework.

\subsection{Data Collection}
\begin{figure}[t]
\begin{center}
\includegraphics[width=\columnwidth]{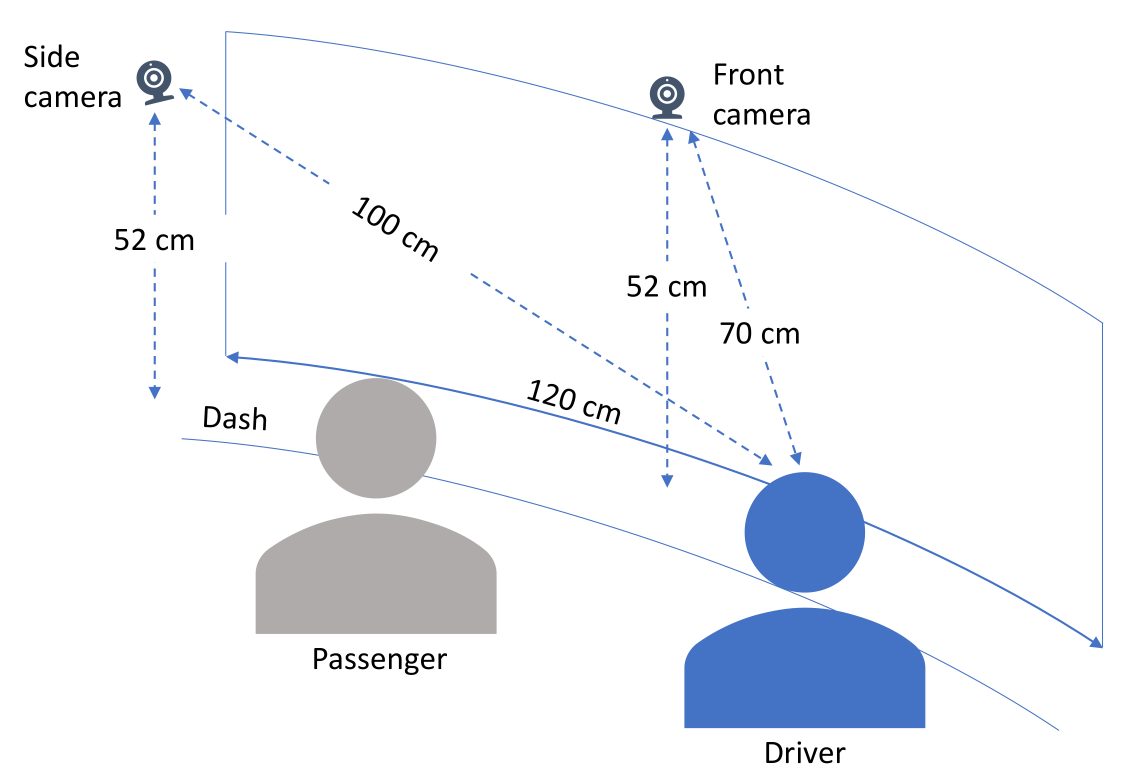}
\end{center}
\vspace*{-0.5em}
\caption{\label{fig:env} Illustration of mock-up car cockpit environment used for data collection. Curved screen shows a driving video to the participant, while the side and front cameras record their body movements and facial expressions, respectively.}
\vspace*{-0.75em}
\end{figure}
We set up the mock-up car cockpit environment sketched in Figure~\ref{fig:env}, which resembles the typical dimensions of a Dailmer AG Smart Fortwo car, and conduct a user behavior study, whereby we emulate automated driving conditions and record data from two cameras (one to the side of the driver, the other front-facing) that capture driver's actions. The mock-up environment follows a setting similar to that described in \cite{taamneh2017multimodal}, but is much simpler, with input data collected only from two cameras and no sensors that are typically used to measure emotional/cognitive stress, as our focus is on routine driver actions with lesser mental load. Both cameras are fixed in terms of locations and always oriented identically for all drivers. The side camera is placed at approximately 1m from the diver's head (Figure~\ref{fig:env}), which allows capturing well body movements. The video streams captured by each camera are time stamped, and the time stamps are aligned. Matching or splicing are thus not required. We recruit a total of 50 participants, 72\% male and 38\% female, with different age, physical appearance, style of clothing, years of driving experience, first spoken language, and level of computer literacy. 

During the experiments, we use a 49-inch Ultra High Definition monitor, onto which each participant is shown 35-minute of 4K dashcam footage of both motorway and urban driving, while being asked to alternate between `driving' and performing a range of tasks. \rev{While we acknowledge that car windshields are inclined and have a trapezoid shape, the driver's field-of-view is largely rectangular and our screen has the same width as the base of the windshield of the representative car we consider. The driving footage was recorded in the UK, hence in our set-up the driver sits on the right hand-side.}

The cameras record the behavior of the participants from different angles, capturing body movements and facial expressions (with a 2$\times$ zoom factor) with 640$\times$360 per-frame pixel resolution and frame rate ranging between 17.14 and 24.74 frames per second (FPS).

\begin{figure}[t]
\begin{center}
\includegraphics[width=\columnwidth]{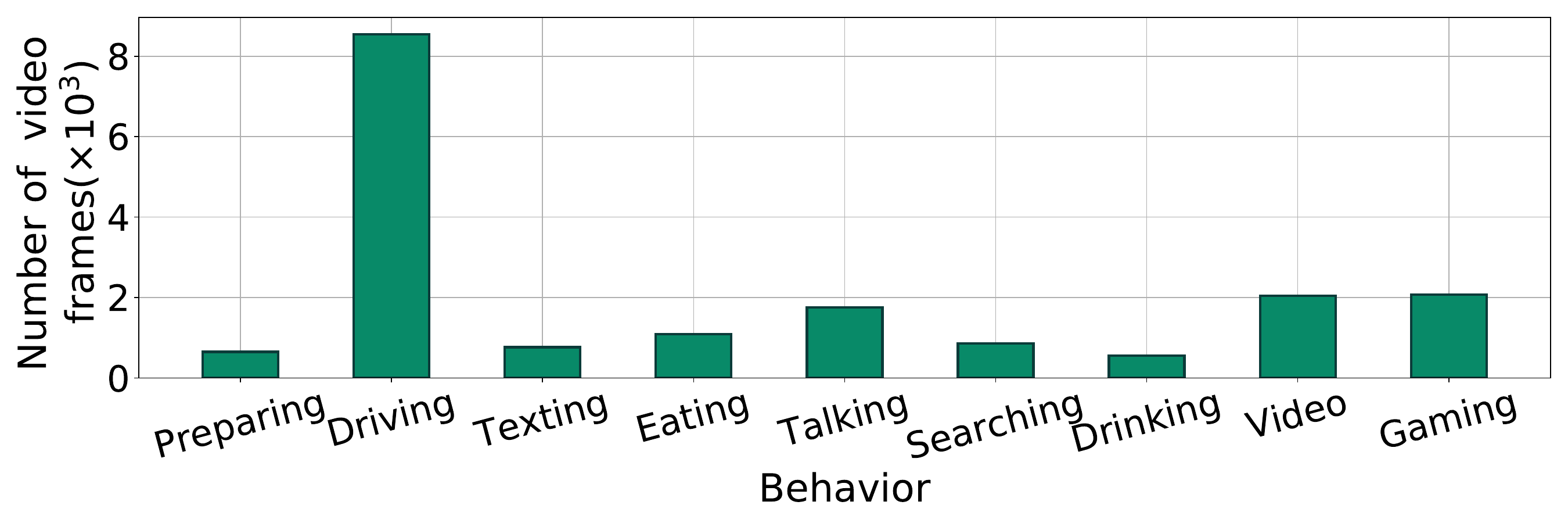}
\end{center}
\caption{\label{fig:count} Summary of the total amount of data (video frames) collected for each driver behavior.}
\end{figure}

We use the OpenCV vision library~\cite{opencv_library} together with Python, in order to label each frame of the approximately 60 hours video recorded, distinguishing between the following actions:
\vspace*{-0.25em}
\begin{enumerate}
    \item \textbf{Normal driving:} The participant focuses on the road conditions shown on the screen and acts as if driving. \rev{This includes both predominant situations with the vehicle in motion and instances where it is stopped at traffic lights, when the driver's hand posture may change.}
    \item \textbf{Texting:} The participant uses a mobile phone to text messages to a friend.
    \item \textbf{Eating:} The participant eats a snack.
    \item \textbf{Talking:} The participant is engaged in a conversation with a passenger.\footnote{We confine consideration to conversations with front-seat passengers, given that average car occupancy is widely below two~\cite{greencongress} and our environment is largely representative of hatchback city cars.}
    \item \textbf{Searching:} The participant is using a mobile phone to find information on the web through a search engine.
    \item \textbf{Drinking:} The participant consumes a soft drink.
    \item \textbf{Watching video:} The participant uses a mobile phone to watch a video.
    \item \textbf{Gaming:} The participant uses a mobile phone to play a video game.
    \item \textbf{Preparing:} The participant gets ready to begin driving or finishes driving.
\end{enumerate}
\vspace*{-0.25em}
The first and last aside, these actions are the most frequently encountered instances of distracted behavior during driving, and are also within the focus of existing datasets (e.g., \cite{abouelnaga2017real, kaggle}). We will consider additional actions that are specific to fully autonomous cars, e.g., sleeping (which would require more sophisticate sensing infrastructure for gaze detection~\cite{Naqvi2018}) as parr of future work. In each experiment, the participant was asked to perform actions (2)--(8) once, while we acknowledge that in real-life driving such behaviors can occur repeatedly. We note that other actions could occur during real driving. However, given the controlled nature of our experiments, we limit inferences to the above well-defined set. Figure~\ref{fig:count} summarizes the amount of data (number of video frames) collected for each type of action that the driver performed. We label the dataset in a  semi-automatic manner. Given a video recorded, we input the start and end time of each behavior, after which the OpenCV tool can label the corresponding frames within that time range automatically based on the FPS rate. This significantly reduces the effort required compared to manual frame-by-frame labeling.\footnote{The dataset collected and source code of our implementation can be accessed via \url{https://github.com/vyokky/interwoven-CNN}.}

\subsection{Data Pre-processing}
\begin{figure}[t]
\begin{center}
\includegraphics[width=\columnwidth]{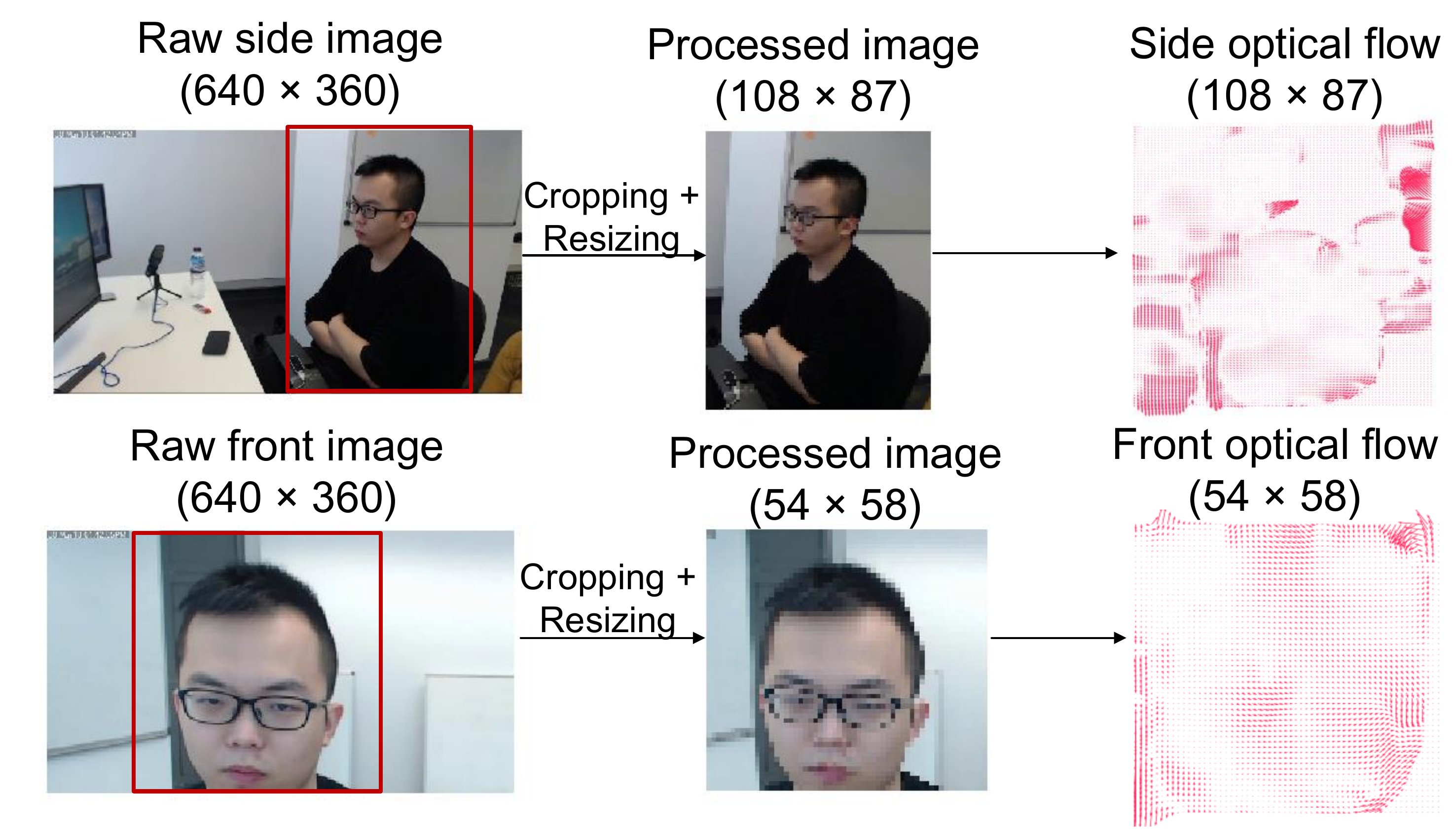}
\end{center}
\caption{\label{fig:process} The data pre-processing procedure (cropping and resizing) and optical flow quiver plots.}
\end{figure}

Recall that the raw videos recorded have 640$\times$360 resolution. Using high-resolution images inevitably introduces storage, computational, and data transmission overheads, which would complicate the model design. Therefore, we employ fixed bounding boxes to crop all videos, in order to remove the background, and subsequently re-size the videos to obtain lower resolution versions. The additional advantage of this approach is that it reduces the adversarial impact of background and illumination changes (as it can often be the case in real driving conditions) on the classification task. Note that the shape and position of the bounding boxes adopted differ between the videos recorded with side and front cameras. We illustrate this process in Figure~\ref{fig:process}.

\begin{figure*}[t]
\centering
\setcounter{subfigure}{0}
\subfigure[The architecture of the Interwoven CNNs (InterCNNs).]{
\label{fig:intercnn} 
\includegraphics[width=\textwidth]{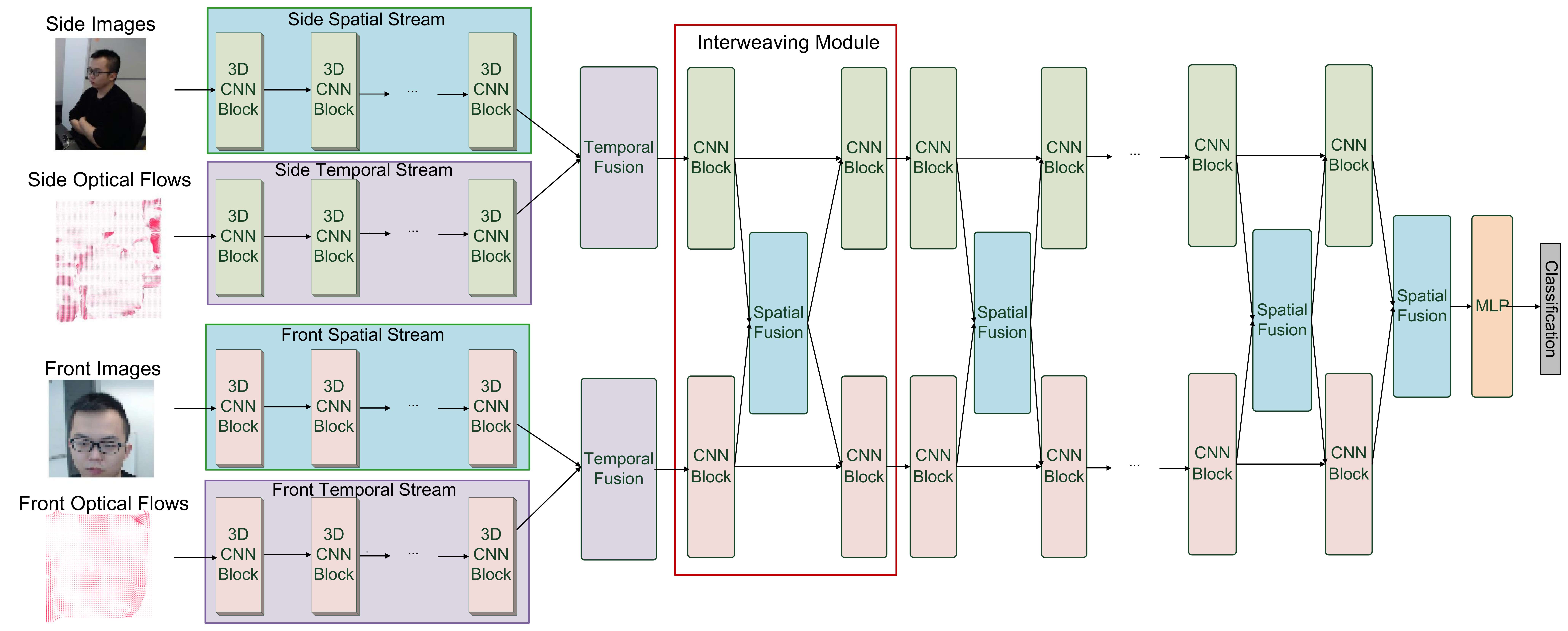}}
\subfigure[The architecture of a plain CNN.]{
\label{fig:cnn} 
\includegraphics[width=0.49\textwidth, valign=top]{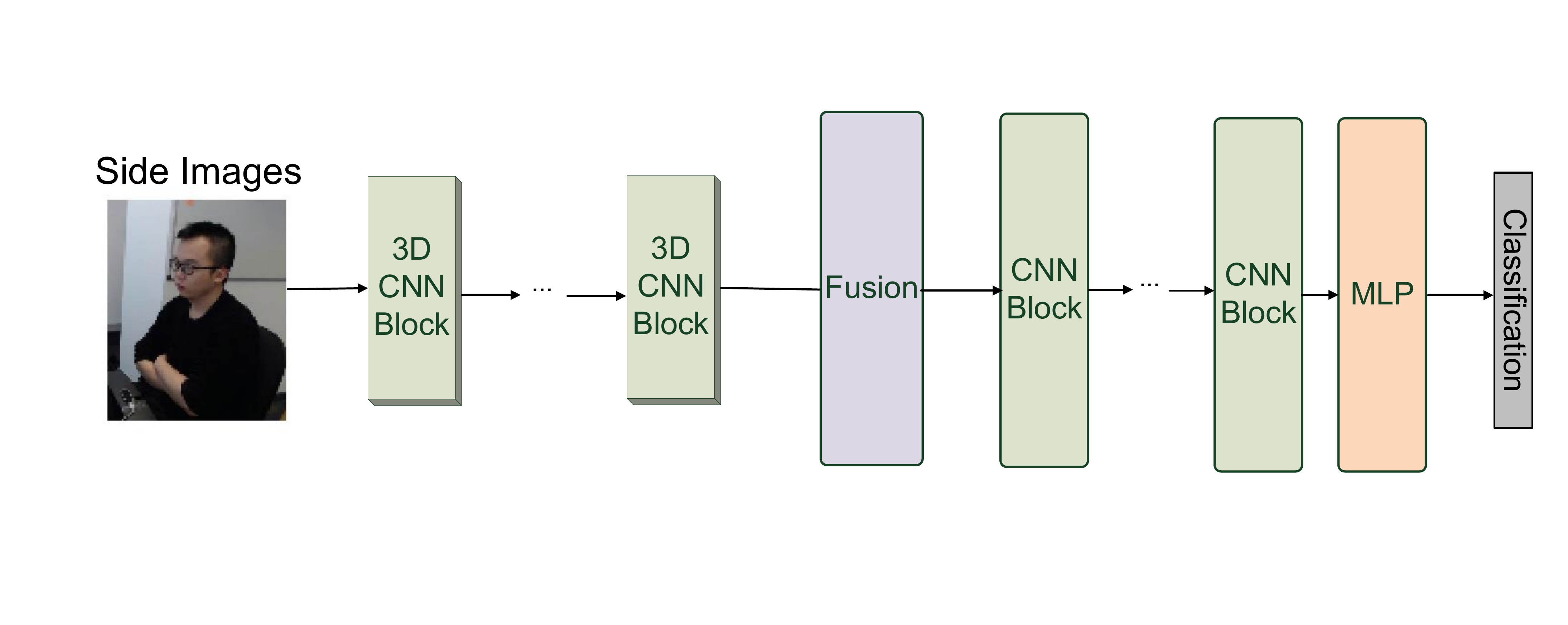}}
\subfigure[The architecture of a two-stream CNN (TS-CNN).]{
\label{fig:cnn_2s} 
\includegraphics[width=0.49\textwidth]{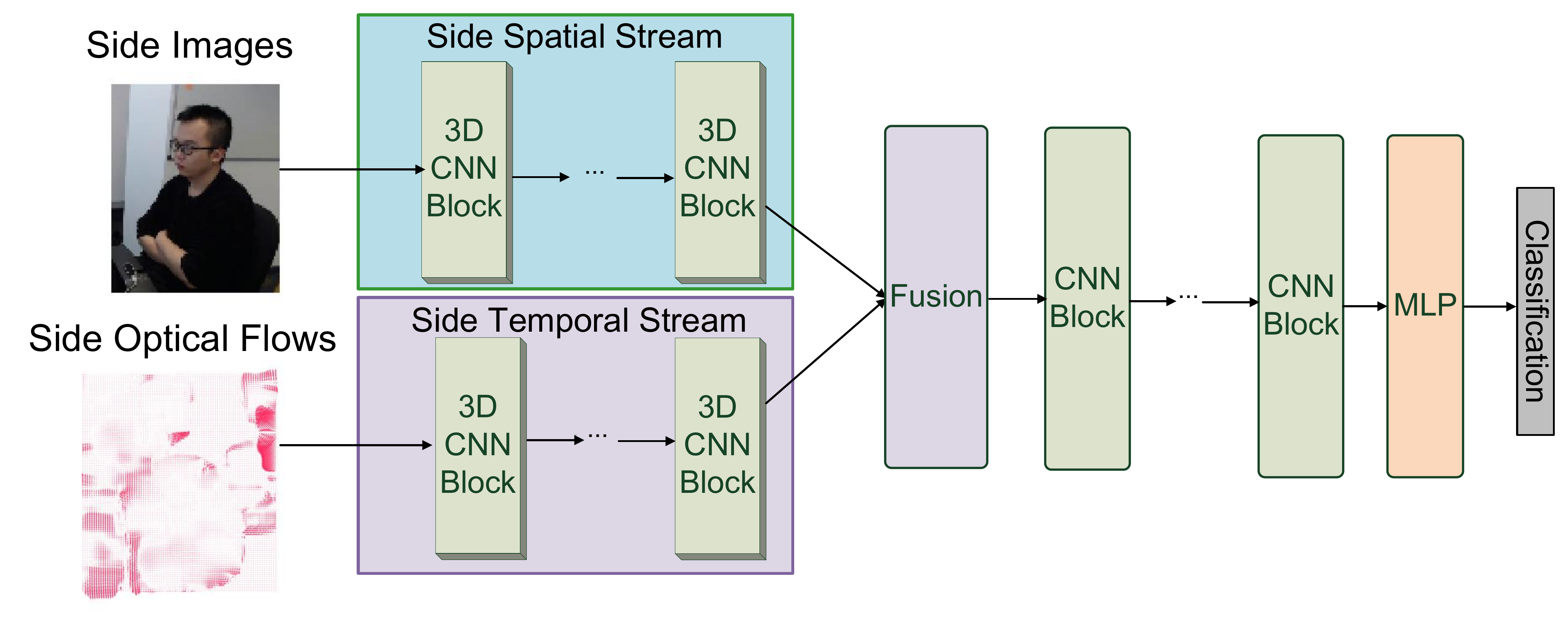}}
\caption{\label{fig:architecture} Three different neural network architectures employed in this study. The plain CNN only uses the side video stream as input, the two-streaming CNN adds extra side OFs, while the InterCNN employs both side and front video streams and optical flows.}
\end{figure*}

Adding Optical Flow (OF) \cite{horn1981determining} to the input of a model has proven effective in improving accuracy \cite{simonyan2014two}. The OF is the instantaneous velocity of the moving objects under scene surface. It can reflect the relationship between the previous and current frames, by computing the changes of the pixel values between adjacent frames in a sequence. Therefore, OF can explicitly describe the short-term motion of the driver, without requiring the model to learn about it. The OF vector $\mathbf{d}^{(x, y)}$ at point $(x, y)$ can be decomposed into vertical and horizontal components, i.e., $\mathbf{d}^{(x, y)} = \{d^{(x, y)}_v, d^{(x, y)}_h\}$. It has the same resolution as the original images, as the computation is done pixel-by-pixel. We show an OF example in Figure~\ref{fig:process}. Our classifier will use OF information jointly with labelled video frames as the input. The experiments we report in Sec.~\ref{Sec:exp} confirm that indeed this leads to superior inference accuracy.

Lastly, we downsample the videos collected, storing only every third frame and obtaining a dataset with 5.71--8.25 FPS, which reduces data redundancy. We will feed the designed model with 15 consecutive video frames and corresponding 14 OF vectors, spanning 1.82 to 2.62 seconds of recording. Such duration has been proven sufficient to capture entire actions, while obtaining satisfactory accuracy \cite{schindler2008action}.

\section{Multi-stream Interwoven CNNs}

We propose a deep neural network architecture specifically designed for classifying driver behavior in real-time, which we call Interwoven Deep Convolutional Neural Network (InterCNN). Our solution uses multi-stream inputs (i.e., side video streams, side optical flows, front video streams, and front optical flows) and merges via fusion layers abstract features extracted by different convolutional blocks. Overall this ensemble demonstrably improves model robustness. We illustrate the overall architecture of our model in Figure~\ref{fig:intercnn}. For completeness, we  also show two simpler benchmark architectures, namely \textit{(i)} a plain CNN, which uses only the side video stream as input (see Figure~\ref{fig:cnn}) and \textit{(ii)} a two-stream CNN (TS-CNN), which takes the side video stream and the side optical flow as input (see Figure~\ref{fig:cnn_2s}). Both of these structures can also be viewed as components of the InterCNN.

\subsection{The InterCNN Architecture}
\label{sec:model}
\begin{figure}[t]
\begin{center}
\includegraphics[width=\columnwidth, trim=15 15 15 15, clip]{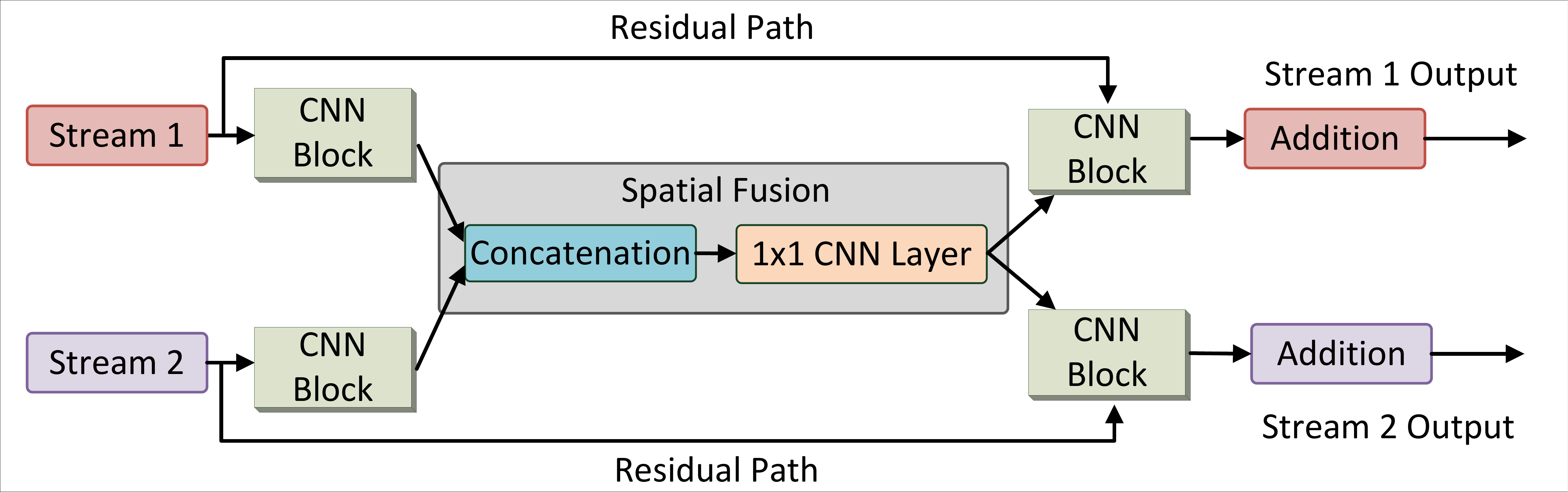}
\end{center}
\caption{\label{fig:inter} Anatomic structure of an Interweaving module.}
\end{figure}

Diving into Figure~\ref{fig:intercnn}, the InterCNN is a hierarchical architecture which embraces multiple types of blocks and modules. It takes four different streams as input, namely side video stream, side optical flow, front video stream and front optical flow. Note that these streams are all four-dimensional (i.e., time, height, width, and RGB channel) for each video frame, and (time, height, width, vertical and horizontal components) for OF frames. The raw data is individually processed in parallel by 7 stacks of 3D CNN blocks. A 3D CNN block is comprised of a 3D convolutional layer to extract spatio-temporal features \cite{ji20133d}, a Batch Normalization (BN) layer for training acceleration \cite{ioffe2015batch}, and a Scaled Exponential Linear Unit (SELU) activation function to improve the model non-linearity and representability \cite{klambauer2017self}. Here, 
$$\textbf{SELU}(\mathbf{x})=\lambda
\begin{cases}
\mathbf{x},& \text{if}\:  \mathbf{x}>0;\\
\alpha e^{\mathbf{x}}-\alpha, & \text{if}\:  \mathbf{x}\leq 0,
\end{cases}$$ 
where the parameters $\lambda=1.0507$ and $\alpha= 1.6733$  are frequently used. We refer to these four streams of 3D CNNs as side spatial stream, side temporal stream, front spatial stream, and front temporal stream respectively, according to the type of input handled. Their outputs are passed to two temporal fusion layers to absorb the time dimension and perform the concatenation operation along the channel axis. Through these temporal fusion layers, intermediate outputs of spatial and temporal streams are merged and subsequently delivered to 25 stacks of Interweaving modules. We illustrate the construction of such modules in Figure~\ref{fig:inter} and detail their operation next.     
\subsection{Interweaving Modules}
The Interweaving module draws inspiration from ResNets \cite{he2016deep} and can be viewed as a multi-stream version of deep residual learning. The two inputs of the module are processed by different CNN blocks individually, and subsequently delivered to a spatial fusion layer for feature aggregation. The spatial fusion layer comprises a concatenation operation and a 1$\times$1 convolutional layer, which can reinforce and tighten the overall architecture, and improve the robustness of the model~\cite{zhang2018long}. Experiments will demonstrate that this enables the model to maintain high accuracy even if the front camera is blocked completely. After the fusion, another two CNN blocks will decompose the merged features in parallel into two-stream outputs again. This maintains the information flow intact.  Finally, the residual paths connect the inputs and the outputs of the final CNN blocks, which facilitates fast backpropagation of the gradients during model training. These paths also build ensembling structures with different depths, which have been proven effective in improving inference accuracy \cite{veit2016residual}. After processing by the Interweaving blocks, the intermediate outputs obtained are sent to a Multi-Layer Perceptron (MLP) to perform the final classification. Notably, the InterCNNs can be trained in an end-to-end manner by backpropagation, that is the weights of the model are adjusted by computing the gradient of the training loss function with different inputs~\cite{Goodfellow-et-al-2016}.

\subsection{CNN Blocks Employed}
\begin{figure*}[t]
\begin{center}
\includegraphics[width=\textwidth, trim=10 10 10 10, clip]{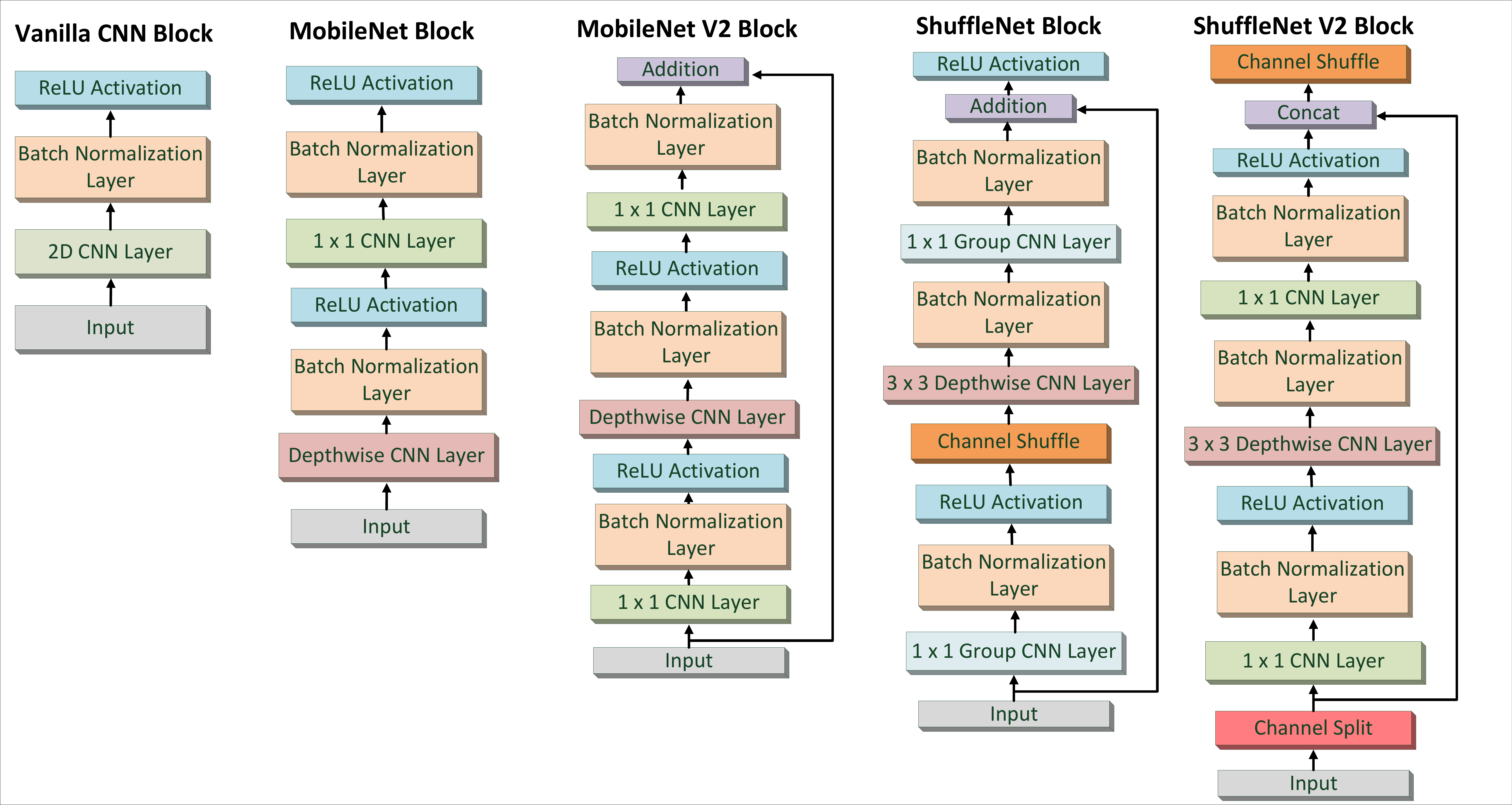}
\end{center}
\caption{\label{fig:cnn_block} \rev{The different CNN blocks employed in this study.}}
\vspace*{-1em}
\end{figure*}

The CNN blocks employed within the interweaving modules are key to performance, both in terms of accuracy and inference time. Our architecture is sufficiently flexible to allow different choices for these CNN blocks. In this work, we explore the vanilla CNN block, MobileNet \cite{howard2017mobilenets} and MobileNet V2 \cite{sandler2018mobilenetv2}, as well as ShuffleNet~\cite{zhang2018shufflenet} and ShuffleNet V2 \cite{ma2018shufflenet} structures, and compare their performance. We are particularly interested in the Mobile-/Shuffle-Net architectures, due to their purpose-built design aimed at execution on embedded devices with limited computation capabilities, such as automotive IoT. We show the architectures of these choices in Figure~\ref{fig:cnn_block} and summarize their parameters in Table~\ref{tab:params}.

\begin{table}[]
    \centering
    \begin{tabular}{c|L{5cm}}
         \textbf{Block}        &  Configuration\\ \hline
CNN    &  36 channels, 3$\times$3 filters, stride$=1$ \\ \hline
MobileNet    &  36 channels, 3$\times$3 filters for Depthwise CNN layer, stride$=1$\\ \hline
MobileNet V2 &  36 channels, 3$\times$3 filters for Depthwise CNN layer, stride$=1$, expand ratio$=6$\\ \hline
ShuffleNet  & 36 channels, 3$\times$3 filters for Depthwise CNN layer, stride$=1$\\ \hline
ShuffleNet V2 & 36 channels, 3$\times$3 filters for Depthwise CNN layer, stride$=1$ \\ \hline
    \end{tabular}
    \caption{Configuration of the different convolutional blocks considered.}
    \label{tab:params}
\end{table}

The vanilla CNN block embraces a standard 2D CNN layer, a BN layer and a Rectified Linear Unit (ReLU) activation function. This is a popular configuration and has been employed in many successful classification architectures, such as ResNet~\cite{he2016deep}. However, the operations performed in a CNN layer are complex and involve a large number of parameters. This may not satisfy the resource constraints imposed by vehicular systems. The MobileNet \cite{howard2017mobilenets} decomposes the traditional CNN layer into a depthwise convolution and a pointwise convolution, which significantly reduces the number of parameters required. Specifically, depthwise convolution employs single a convolutional filter to perform computations on individual input channels. In essence, this generates an intermediate output that has the same number of channels as the input. The outputs are subsequently passed to a pointwise convolution module, which applies a 1$\times$1 filter to perform channel combination. MobileNet further employs a hyperparameter $\alpha$ to control the number of channels, and $\rho$ to control the shape of the feature maps. We set both $\alpha$ and $\rho$ to 1 in our design. In summary, the MobileNet block breaks the filtering and combining operations of a traditional CNN block into two layers. This significantly reduces the computational complexity and number of parameters required, while improving efficiency in resource-constrained devices.

The MobileNet V2 structure \cite{sandler2018mobilenetv2} improves the MobileNet by introducing an inverted residual and linear bottleneck. The inverted residual incorporates the residual learning specific to ResNets. The input channels are expanded through the first 1$\times$1 convolutional layer, and compressed through the depthwise layer. The expansion is controlled by a parameter $t$, which we set to 6 as default. To reduce information loss, the ReLU activation function in the last layer is removed. Compared to MobileNet, the second version has fewer parameters and higher memory efficiency. As we will show, this architecture may sometimes exhibit superior accuracy. Both MobileNet and MobileNet V2 are tailored to embedded devices, as they make inferences faster with smaller models. These makes them suitable for in-vehicle classification systems.

ShuffleNet employs pointwise group convolution and channel shuffle to reduce computation complexity, so as to meet the constraints of mobile platforms, while maintaining accuracy~\cite{zhang2018shufflenet}. Channel shuffling is a differentiable operation that enables information to  flow between different groups of channels, thereby improving representation while allowing to train the stucture end-to-end. ShuffleNet V2 improves the efficiency of its precursor architecture, by introducing a channel split operator~\cite{ma2018shufflenet}. This splits the input feature channels and reduces the complexity associated with group convolution operations by replacing these with regular 1$\times$1 convolutions and branch concatenation, with channel shuffle allowing information exchanges between the branches. In the same spirit as the MobileNet blocks, ShuffleNet (V2) should be particularly well-suited to behavior classification on embedded platforms, yet the performance difference between the two on this task remains to be uncovered.

\begin{figure}[t]
\begin{center}
\includegraphics[width=0.5\textwidth]{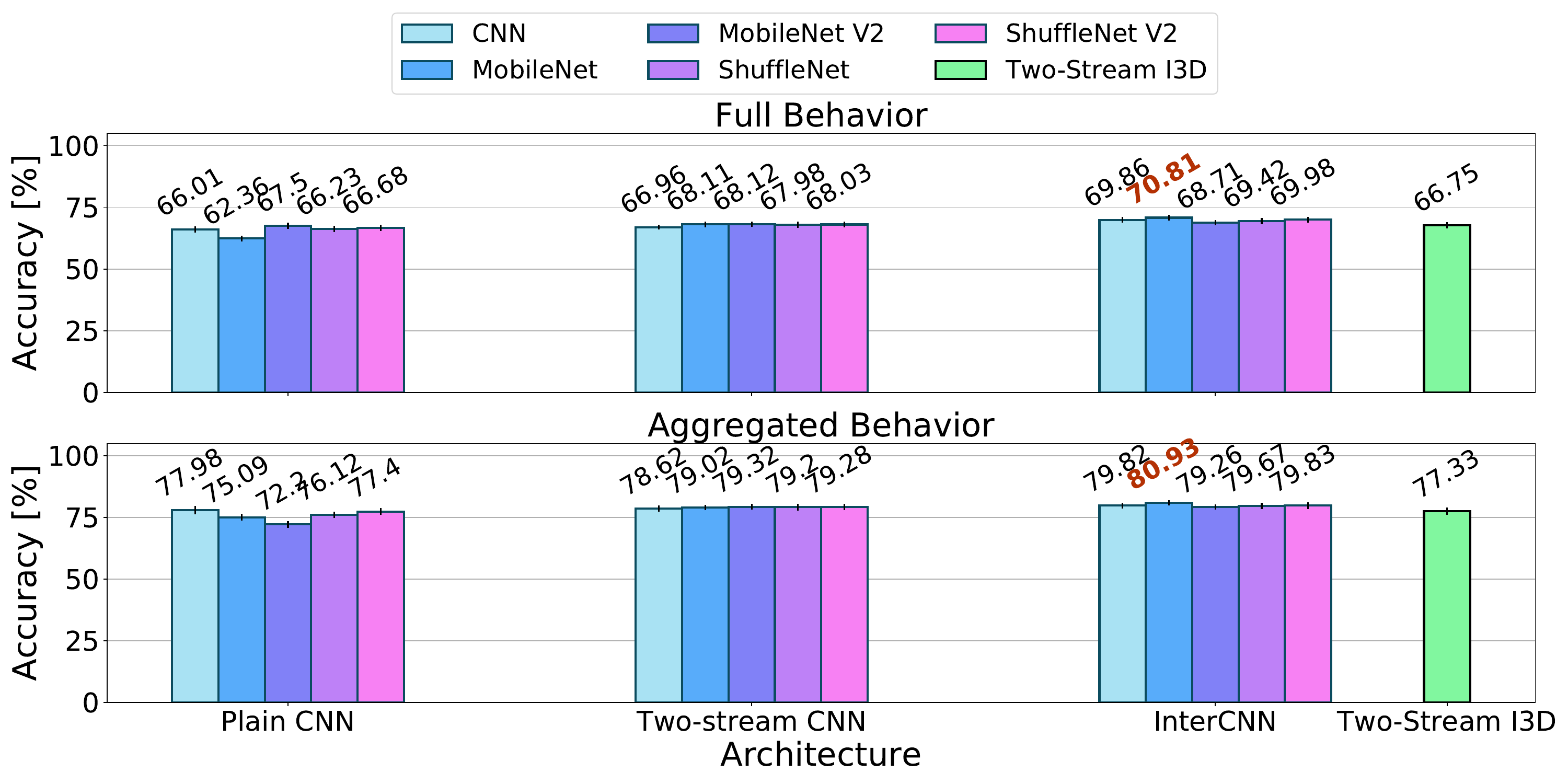}
\end{center}
\caption{\label{fig:acc} Prediction accuracy in the case of classification among 9 different driver behaviors (top) and aggregate tasks (bottom), for all the neural network architectures considered.}
\end{figure}

\section{Experiments}\label{Sec:exp}
In this section, we first describe briefly the implementation of the proposed InterCNN for driver behavior recognition, then compare the prediction accuracy of different CNN blocks that we can incorporate in this architecture. Subsequently, we examine complexity--accuracy tradeoffs, introduce a temporal voting scheme to improve performance, and show that our architecture is robust to losses in the input. Finally, we dive deeper into the operation of the proposed model, visualizing the output of the hidden layers, which helps understanding what knowledge is learned by the InterCNN.

\subsection{Implementation}
We implement the different neural network architectures studied in TensorFlow version 1.4.0 \cite{tensorflow} with the TensorLayer version 1.6.7 library~\cite{tensorlayer}.
We train all models using a batch size of 100, on average for 136 epochs, each with 10,600 iterations, using the Adam optimization algorithm \cite{kingma2015adam}, which is based on stochastic gradient descent. With this we seek to minimize the standard cross-entropy loss function between true labels and the outputs produced by the deep neural networks.
Training is performed for approximately 10 days (with early-stop based on the validation error) on a computing cluster with 18 nodes, each equipped with two Intel Xeon E5-2620 processors (24 logical cores) clocked at 2.1 GHz, 64 GB of RAM and a mix of multiple NVIDIA TITAN X and Tesla K40M GPUs, each with 12GB of memory. To maintain consistency when evaluating their computation efficiency, we test all models using a workstation equipped with an Intel Xeon W-2125 CPU clocked at 2.66 GHz, 32GB of RAM, and an NVIDIA TITAN X GPU.

\subsection{Accuracy Assessment}
We randomly partition the entire dataset into a training set (30 videos), a validation set (10 videos), and a test set (10 videos), and repeat the partitioning process 5 times, reporting average performance across 5 different test sets for completeness. We assess the accuracy of our solution on two categories of behaviors. The first considers all the 9 different actions performed by the driver (see Sec.~\ref{sec:data}). In the second round, we aggregate the behaviors that are visually similar and carry similar user cognitive status. In particular, [Texting, Searching, Watching Video, Gaming] are aggregated into a ``Using phone'' behavior, and [Eating, Drinking] are combined into a single ``Eat \& Drink'' action. To put things into perspective, we also consider the two stream inflated 3D CNN (I3D) \cite{valeriano2018recognition} as a baseline approach.

In Figure~\ref{fig:acc}, we show the prediction accuracy of the InterCNN architecture with all the CNN block types considered, as well as that of plain and two-stream CNN architectures, each employing the same three types of blocks. Observe that in the case of ``full behavior'' recognition (top subfigure), the proposed \mbox{Inter}CNN with MobileNet blocks achieves the highest prediction accuracy, outperforming the plain CNN by 8.45\%. Further, we can see that \textbf{feeding the neural network with richer information (optical flows and front images) improves accuracy, as our two-stream CNN and InterCNN on average outperform the plain CNN by 8.44\% and 4.50\% respectively}. This confirms that the OF and facial expressions provide useful descriptions of the behaviors, which our architectures effectively exploits. Both structures further outperform the baseline two-stream I3D. It is also worth noting that, although the performance gains over the plain CNN may appear relatively small, the amount of computational resource required by our architecture, inference times, and complexity are significantly smaller. We detail these aspects in the following subsection.

Turning attention to the aggregated behavior (bottom subfigure), observe that the accuracy improves significantly compared to when considering all the different actions the driver might perform, as we expect. This is because some behaviors  demonstrate similar patterns (e.g., texting and web searching), which makes discriminating among these extremely challenging. Overall, \textbf{the InterCNN with MobileNet blocks obtains the best prediction performance when similar behaviors are aggregated, outperforming other architectures by up to 5.84\%}. In addition, our two-stream CNN and InterCNN consistently outperform the plain CNN, as well as the existing two-stream I3D that we consider as baseline.

\subsection{Model Complexity \& Inference Time}\label{sec:com}
Next, we compare the model size (in terms of number or weights and biases to be configured in the model), inference time, and complexity (quantified through floating point operations -- FLOPs) of InterCNNs with different CNN blocks. Lower model size will pose smaller storage and memory requirements on the in-vehicle system. The inference time refers to how long it takes to perform one instance of driver behavior recognition. This is essential, as such applications are required to perform in real-time. Lastly, the number of FLOPs \cite{oberman1997design} is computed by counting the number of mathematical operations or assignments that involve floating-point numbers, and is routinely used to evaluate the complexity of a model. 
\begin{figure*}[t]
\begin{center}
\includegraphics[width=0.7\textwidth]{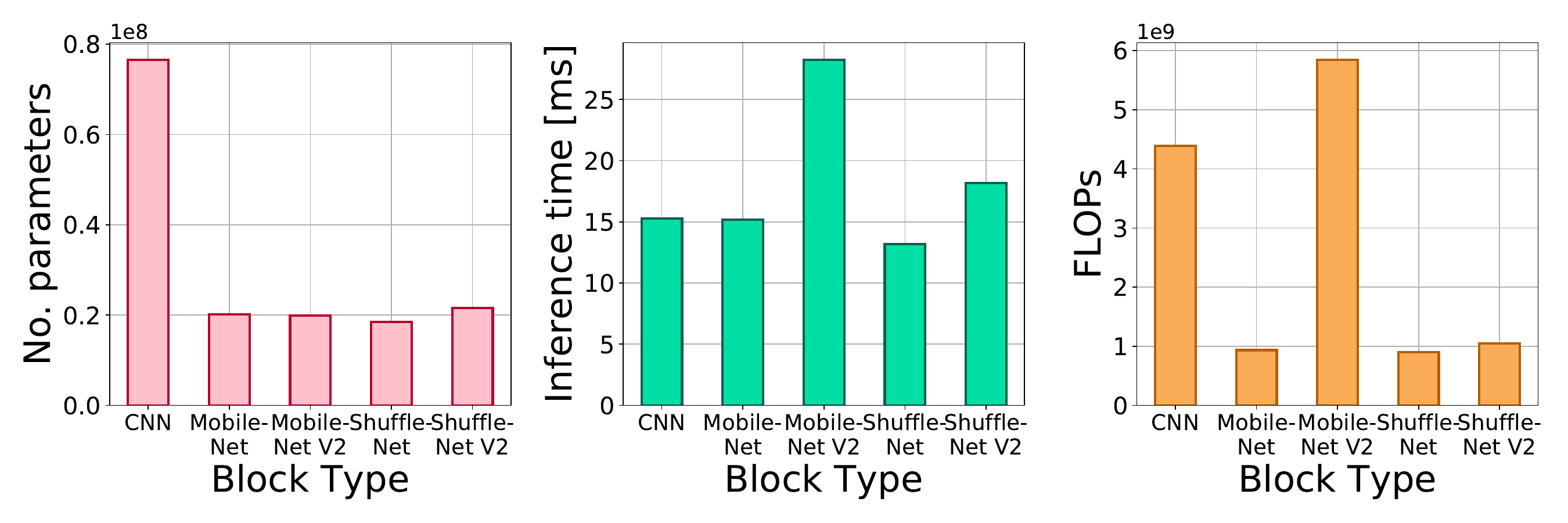}
\end{center}
\caption{\label{fig:efficiency} Comparison of number of parameters (left), inference time (middle) and FLOPS (right) on InterCNNs with different CNN blocks.}
\end{figure*}

We illustrate this comparison in Figure~\ref{fig:efficiency}. Observe that Mobile-/Shuffle-Net (V2) blocks have similar numbers of parameters, and these are 4 times fewer than those of vanilla CNN blocks. This is consistent with the conclusions drawn in \cite{howard2017mobilenets} and \cite{sandler2018mobilenetv2}. Overall, when adopting MobileNet V2 blocks in all the architectures shown in Figure~\ref{fig:architecture}, our Interwoven CNN involves training 19,870,139 parameters; this is only about 8\% more than the number trained with the plain CNN and TS-CNN structures, which have 18,247,277 and respectively 18,299,099  parameters. Observe also that, \textbf{InterCNNs with MobileNet/ShuffleNet blocks can infer driver behavior within 15 ms per instance} (center subfigure) with the help of a GPU, which satisfies the real-time constraints of intelligent vehicle systems.
In addition, our approach require 28 ms, 25 ms and 23 ms when plain CNN, TS-CNN and interwoven CNN are employed with MobileNet V2 blocks.
Note that the computation time is obtained using an NVIDIA Tesla K40M GPU, which can perform 1.682 TFLOPS. Considering the NVIDIA Jetson AGX Xavier in-vehicle GPU,\footnote{Jetson AGX Xavier, \url{https://developer.nvidia.com/embedded/jetson-agx-xavier}} which can deliver 5 TFLOPS, our architectures can easily perform inferences in a real-time manner on embedded systems. Runtime performance is indeed similar to that of an architecture employing CNN blocks, yet less complex, while an architecture with MobileNet blocks is 46.2\% faster than with MobileNet V2 blocks. Lastly, \textbf{the number of FLOPs required by an InterCNN with MobileNet or ShuffleNet (V2) blocks is approximately 4.5 and 6 times smaller than when employing CNN and respectively MobileNet V2 blocks.} This requirement can be easily handled in real-time even by a modern CPU.

\subsection{Temporal Voting}
\begin{figure}[t]
\begin{center}
\includegraphics[width=\columnwidth]{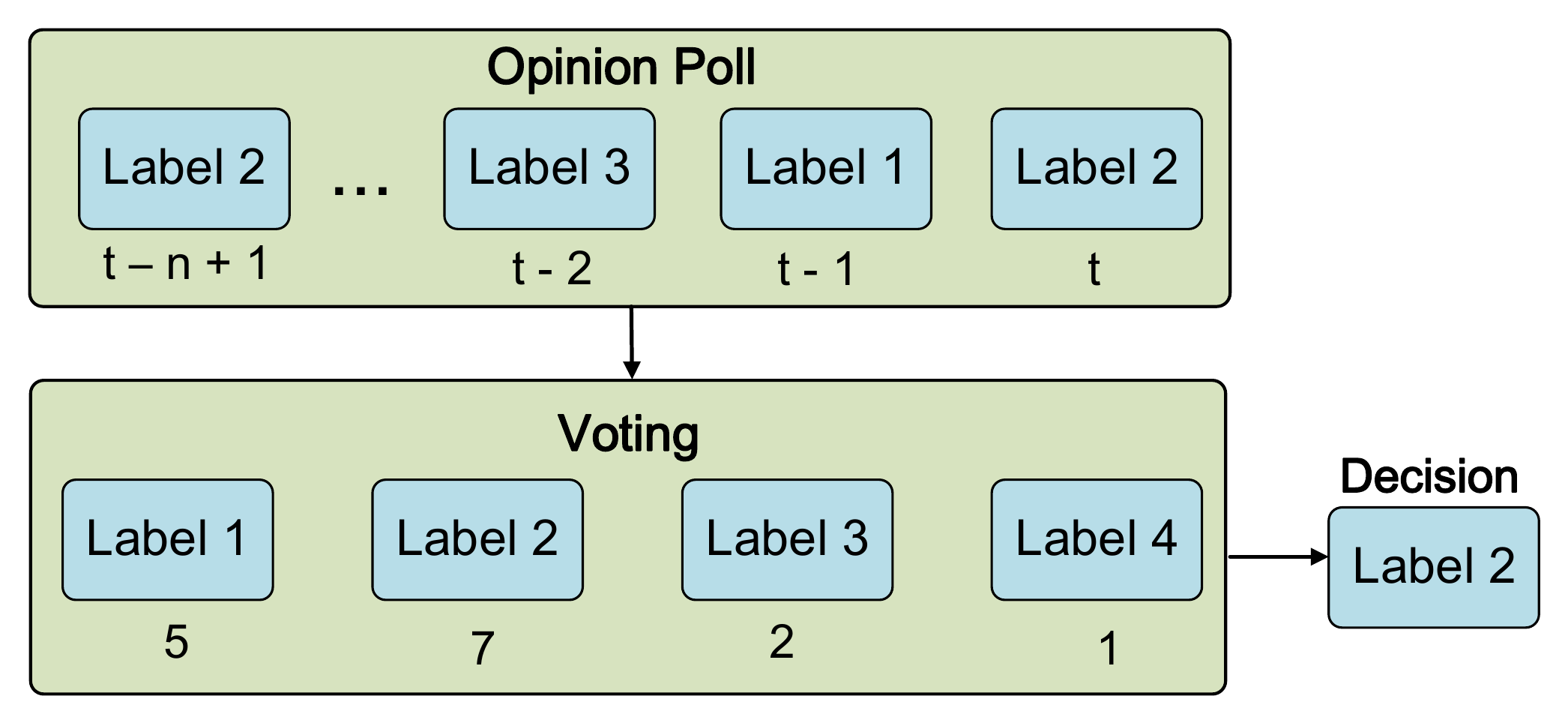}
\end{center}
\caption{\label{fig:tv} Illustration of the Temporal Voting (TV) scheme.}
\end{figure}

In the collected dataset, since the videos are recorded at high FPS rate, we observe that the behavior of a driver will not change very frequently over consecutive frames that span less than 3 seconds. Therefore, we may be able to reduce the likelihood of  misclassification by considering the actions predicted over recent frames. To this end, we employ a temporal voting scheme, which constructs an opinion poll storing the inferred behaviors over the $n$ most recent frames, and executes a ``voting'' procedure. Specifically, the driver's action is determined by the most frequent label in the poll. We illustrate the principle of this Temporal Voting (TV) procedure in Figure~\ref{fig:tv}. We set $n=15$, by which the poll size bears the same temporal length as the inputs. 

We show the prediction accuracy gained after applying TV in Tables~\ref{tab:tv-f} and \ref{tab:tv-a}. Observe that the TV scheme improves the classification accuracy of all architectures on both full and aggregated behavior sets. In particular, on average the accuracy on full behavior recognition increases by 2.17\%, and that of aggregated behavior recognition by 1.78\%. This demonstrates the effectiveness of the proposed TV scheme.
\begin{table}[!h]
\caption{Inference accuracy gained with different CNN blocks over full behaviors, before\textbackslash{}after applying the TV~scheme. \label{tab:tv-f}}
\centering
\begin{tabular}{|l|l|l|l|}
\hline
\textbf{Block}        & Plain CNN                      & TS-CNN                         & InterCNN                       \\ \hline
CNN    & 3.40\% & 1.23\% & 1.67\% \\ \hline
MobileNet    & 1.03\% & 2.45\% & 3.21\% \\ \hline
MobileNet V2 & 1.98\% & 2.46\% & 2.17\% \\ \hline
ShuffleNet  & 2.02\% & 2.41\% & 2.20\% \\ \hline
ShuffleNet V2 & 1.84\% & 2.35\% & 2.09\% \\ \hline
\end{tabular}
\end{table}

\begin{table}[!h]
\caption{Inference accuracy gained with different CNN blocks over aggregated behaviors before\textbackslash{}after applying the TV scheme.\label{tab:tv-a}}
\centering
\begin{tabular}{|l|l|l|l|}
\hline
\textbf{Block}        & Plain CNN                      & TS-CNN                         & InterCNN                       \\ \hline
CNN block    & 1.65\% & 1.83\% & 1.90\% \\ \hline
MobileNet    & 2.52\% & 2.11\% & 0.98\% \\ \hline
MobileNet V2 & 1.88\% & 1.55\% & 1.63\% \\ \hline
ShuffleNet  & 2.23\% & 2.31\% & 2.16\% \\ \hline
ShuffleNet V2 & 1.97\% & 2.19\% & 2.00\% \\ \hline
\end{tabular}
\end{table}

\begin{figure}[htb]
\begin{center}
\includegraphics[width=0.5\textwidth]{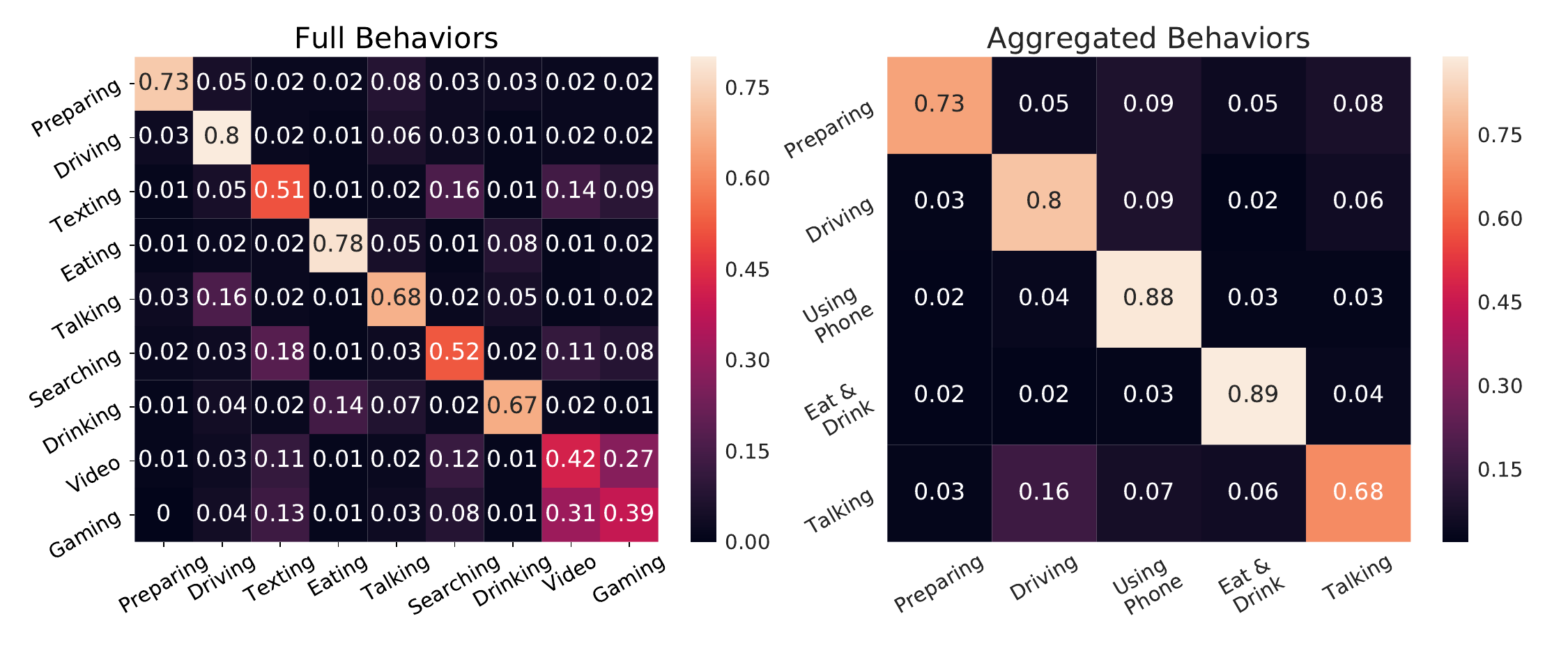}
\end{center}
\caption{\label{fig:cm} The normalized confusion matrices for both classification cases with the TV scheme using InterCNNs.}
\end{figure}

Further, we show normalized confusion matrices in Figure~\ref{fig:cm} for both classification cases, with the TV scheme using InterCNNs. For full behaviors classification, observe that our InterCNNs do not classify the ``watching video'' and ``gaming'' particularly well, as they are visually similar. When aggregating these with ``texting'' and  ``searching'' actions into a ``using phone'' class, the accuracy increases to 89\%. Further, ``talking'' is difficult for the model to classify, since this behavior does not encompass a distinctive feature, such as holding a phone or a drink. Incorporating additional sensors (\emph{e.g.} microphone), may help improve performance. \textbf{Overall, our model performs well on aggregated behaviors, as accuracy improves by 8\%}.

\subsection{Model Insights}
Lastly, we delve into the inner workings of the InterCNN by visualizing the output of the hidden layers of the model, aiming to better understand how the neural network ``thinks'' of the data provided as input and what knowledge it learns.
\begin{figure*}[htb]
\begin{center}
\includegraphics[width=0.9\textwidth]{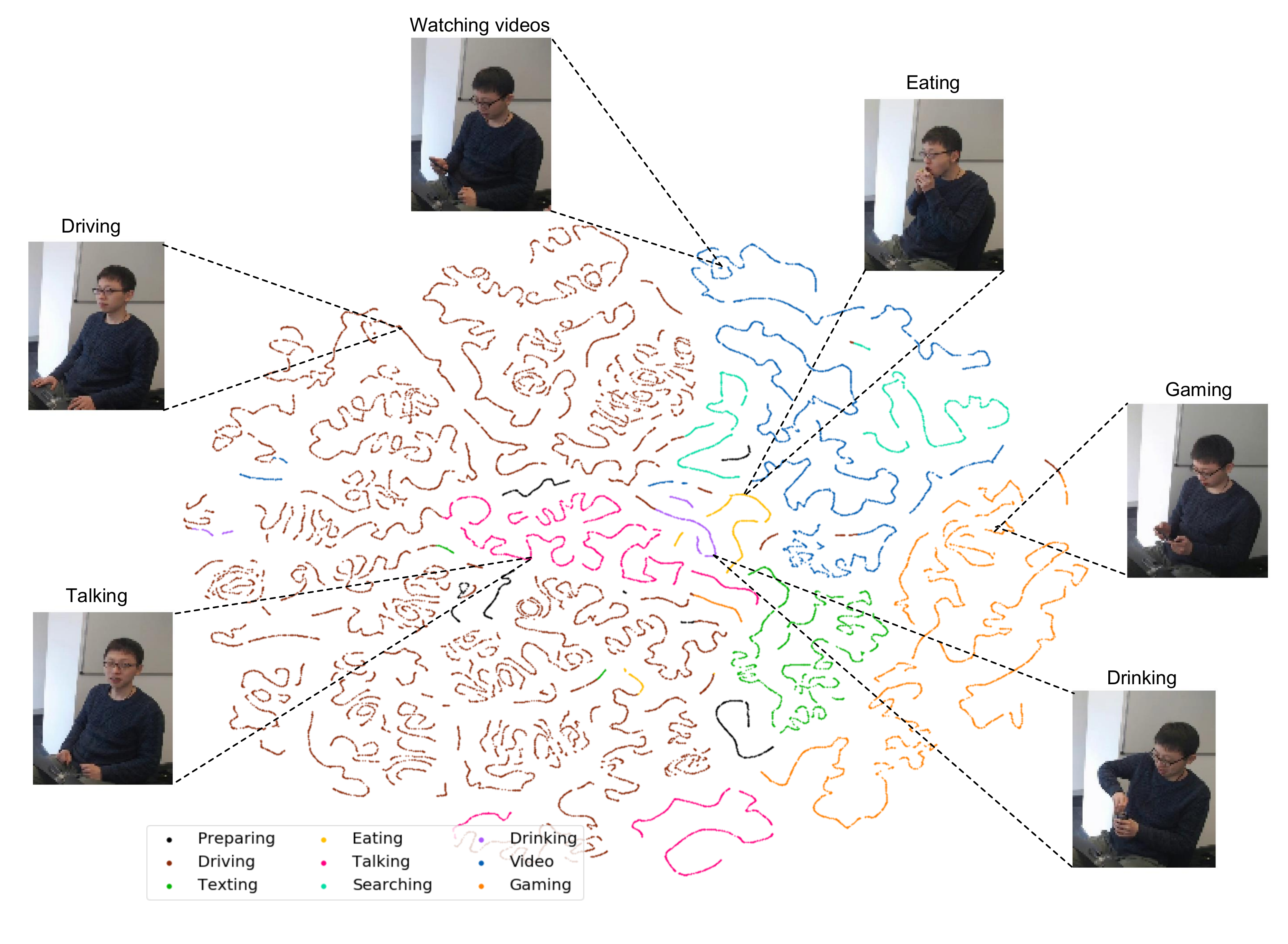}
\end{center}
\caption{\label{fig:tsne} Two-dimensional t-SNE embedding of the representations in the
last hidden layer of the InterCNN with MobileNet blocks. Data generated using a full test video (35,711 frames).}
\end{figure*}

\vspace*{0.5em}
\noindent\textbf{T-distributed Stochastic Neighbor Embedding Vizualization:}
We first adopt the t-distributed Stochastic Neighbor Embedding (t-SNE) \cite{maaten2008visualizing} to reduce the dimension of the last layer (the MLP layer in Figure~\ref{fig:intercnn}), and plot the hidden representations of a testing video (35,711 frames) into a two-dimensional plane, as shown in Figure~\ref{fig:tsne}. In general, the t-SNE approach arranges data points that have a similar code nearby in the embedding.
In general, t-SNE is a nonlinear dimensionality reduction technique that is used to embed high-dimensional data, so as to enable visualization in a low-dimensional (e.g., 2D) space, in terms of pairwise distance between points. This typically reflects how the model ``thinks'' of the data points, as similar data representations will be clustered together.

\begin{figure}[htb]
\begin{center}
\includegraphics[width=0.5\textwidth]{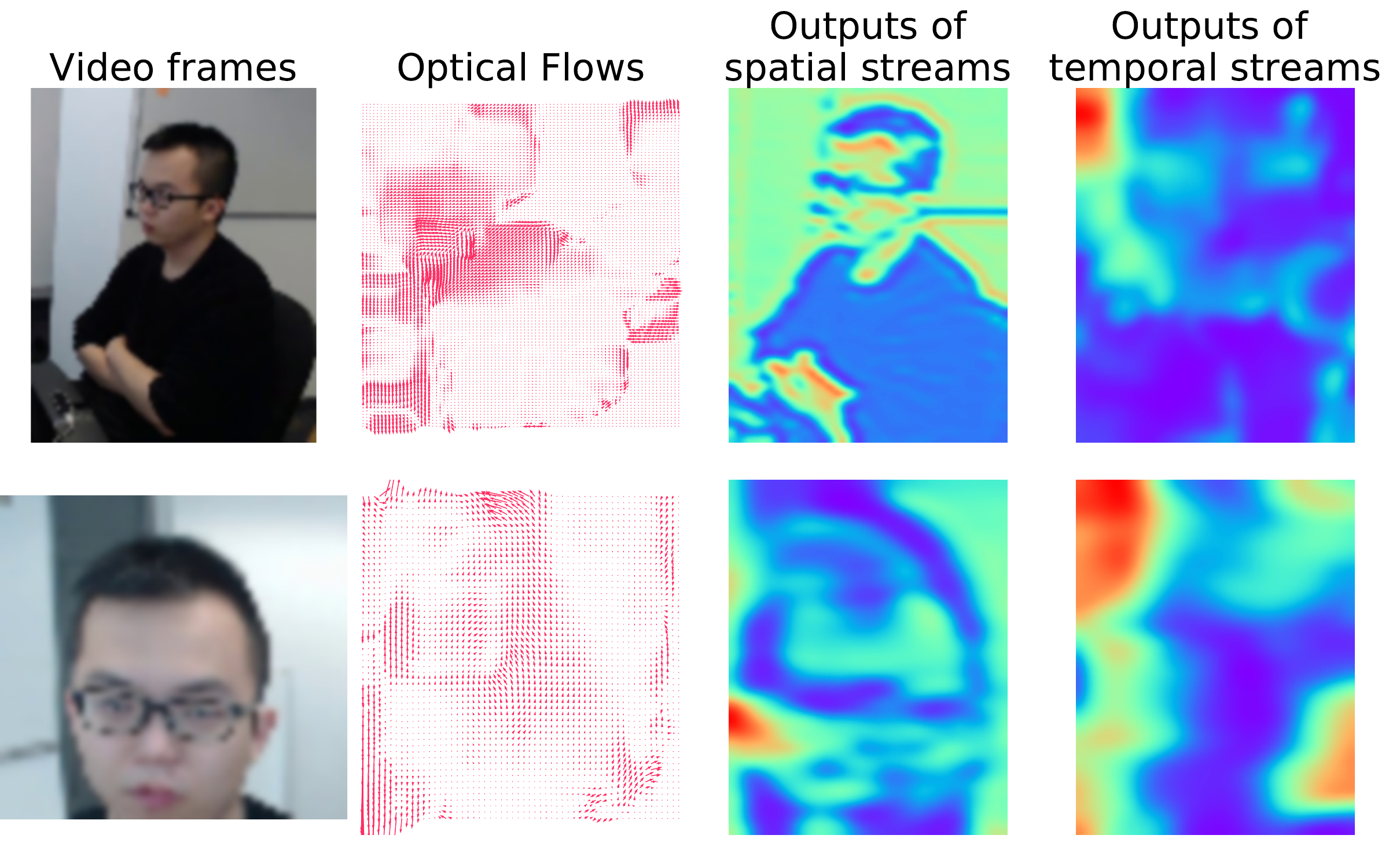}
\end{center}
\caption{\label{fig:hidden} The input videos frames, optical flows, and hidden outputs of the 3D CNN blocks before two temporal fusion layers. Figures shown correspond to a single instance.}
\vspace*{-1.5em}
\end{figure}

Interestingly, the embeddings of the ``Eating'' and ``Drinking'' actions remain close to each other, as both actions require to grasp a snack or drink and bring this close to the mouth. Furthermore, the embeddings of actions that use a phone (i.e., ``Web searching'', ``Texting'', ``Watching videos'', and ``Gaming'') are grouped to the right side of the plane, as they are visually similar and difficult to differentiate. Moreover, as ``Preparing'' involves a combination of actions, including sitting down and talking to the experiment instructor, its representation appears scattered. These observations suggest that our model effectively learns the feature of different behaviors after training, as it projects similar data points onto nearby positions.

\vspace*{0.5em}
\noindent\textbf{Hidden Layer Output Visualization:}
We also investigate the knowledge learned by the model from a different perspective, by visualizing the output of the hidden layers of the 3D CNN block before the temporal fusion layers. This will reflect the features extracted by each individual neural network stream. We show a snapshot of such visualization in Figure~\ref{fig:hidden}. Observe that the spatial streams perform  ``edge detection'', as the object edges in the raw inputs are outlined by the 3D CNN. On the other hand, the output of the hidden layers in the temporal steams, which process the optical flows, are too abstract to interpret. In general, the abstraction level of the features extracted will increase with the depth of the architecture; it is the sum of such increasingly abstract features that enables the neural network to perform the final classification of behaviors with high accuracy.

We conclude that, by employing MobileNet blocks in InterCNNs, we achieve the highest accuracy in the driver behavior recognition task, as compared with any of the other candidate CNN block architectures. The InterCNN + MobileNet combo also demonstrates superior computational efficiency, as it requires the lowest number of parameters, exhibits the fastest inference times and the least FLOPs. Importantly, our design is robust to lossy inputs. The sum of these advantages makes the proposed InterCNN with MobileNet blocks an excellent solution for accurate driver behavior recognition, easily pluggable in modern in-vehicle intelligent systems.

\subsection{Discussion}
Finally, we discuss the advantages of a dual-view approach to driver behavior recognition, as well as limitations and potential improvements of our InterCNN.

\begin{figure}[t]
\begin{center}
\includegraphics[width=0.5\textwidth]{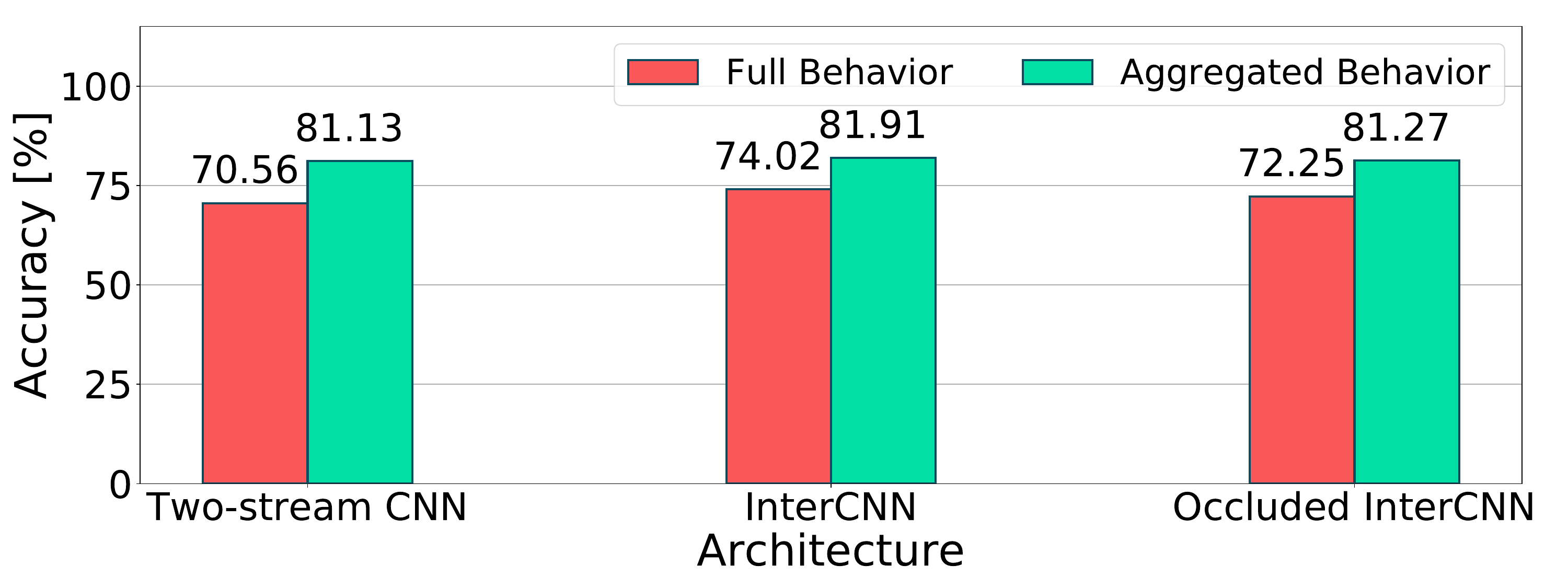}
\end{center}
\caption{\label{fig:block} Comparison of accuracy of two-stream CNN, InterCNN, and InterCNN with occluded inputs. All architectures employ MobileNet blocks.}
\end{figure}

The key benefit of operating with two video streams as is the improvement of the system's robustness to lossy input. To demonstrate this we block the front video and the front OF inputs when performing inferences, and examine again the classification accuracy.
Such circumstances can occur in real world settings, e.g., the camera may be intentionally or accidentally occluded by the driver. To cope with such conditions, we fine-tune our model by performing ``drop-outs'' over the inputs \cite{srivastava2014dropout}. Specifically, we block the front video and the front OF streams with probability $p = 0.5$ during the training and testing of the InterCNN with MobileNet blocks. We summarize the obtained performance in Figure~\ref{fig:block}. Note that by blocking the front video and the front OF streams, the input of the InterCNN is the same as that fed to the two-stream CNN, while the architecture remains unchanged.

Observe that although the prediction accuracy of InterCNN drops slightly when the inputs are blocked, the InterCNN with occluded inputs continues to outperform the two-steam CNN in the full behavior recognition task. This suggests that our proposed architecture with Interweaving modules is effective and highly robust to lossy input. 

\begin{figure}
    \centering
    \includegraphics[width=0.8\columnwidth]{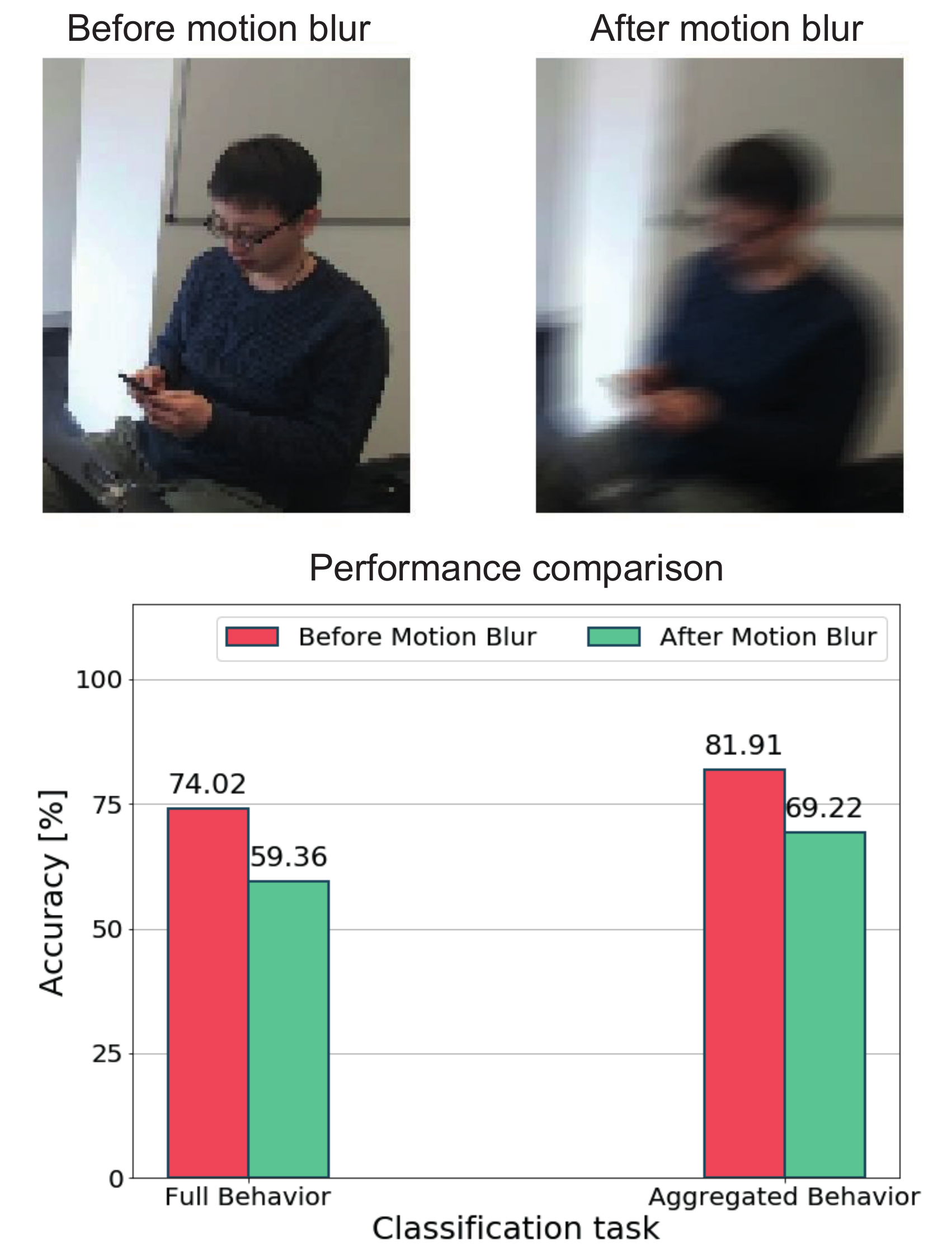}
    \caption{\rev{Sample frames with/without motion blur and classification performance of InterCNN with MobileNet modules in the absence/presence of this effect.}}
    \label{fig:blur}
\end{figure}

\rev{Another aspect that may be important in practical settings where the cameras that provide input to our driver behavior recognition system do not incorporate robust image stabilization software, is motion blur induced by shaking due to uneven road surfaces. To examine the impact of such phenomenon, 
we apply motion blur (degree = 10, angle = 60) to every frame in the test set with 20\% probability, without retraining. The consequences of this effect and the performance assessment of the InterCNN with MobileNet modules, with and without this effect are shown in Figure~\ref{fig:blur}. We notice that `bumpy' roads reduce the classification performance on aggregate behaviors by approximately 15\%. This emphasizes the importance of image stabilization technology, which has become affordable and is already widely incorporated in off-the-shelf cameras.}

While our experiments have been conducted in a tightly controlled environment, further research is needed to consider a richer set of actions, which the driver may conduct (including reaching out for items in the car cockpit or operating car entertainment systems), different driver background settings due to travel, or level of cognitive decline (e.g., tiredness, micro-sleep), which call for more sophisticated biometric/eye-tracking infrastructure. Nonetheless, the proposed InterCNN offers promising results towards achieving practical full vehicle automation.

\section{Conclusions and Future work}
\label{sec:conclusions}
In this paper, we proposed an original Interwoven Convolutional Neural Network (InterCNN) to perform driver behavior recognition. Our architecture can effectively extract information from multi-stream inputs that record the activities performed by drivers from different perspectives (i.e., side video, side optical flow, front video, and front optical flow), and fuse the features extracted to perform precise classification. We further introduced a temporal voting scheme to build an ensembling system, so as to reduce the likelihood of misclassification and improve accuracy. Experiments conducted on a real-world dataset that we collected with 50 participants demonstrate that our proposal can classify 9 types of driver behaviors with 73.97\% accuracy, and 5 classes of aggregated behaviors with 81.66\% accuracy. Our model makes such inferences within 15 ms per instance, which satisfies the real-time constraints of modern in-vehicle systems. The proposed InterCNN is further robust to lossy data, as inference accuracy is largely preserved when the front video and front optical flow inputs are occluded.

Future work will involve collecting data from additional sensors mounted on a real car, such as infrared cameras, breaking, speed, car body sensors and from external sources (e.g. weather reports). We expect this to help build more robust driver behavior recognition systems that can operate in more challenging conditions, such as driving at night or during heavy rain/fog conditions.

 \section*{Acknowledgments}

This research was supported by grant no. 18TLRP-B131486-02 from the Transportation and Logistics R\&D Program funded by the Ministry of Land, Infrastructure and Transport of the Korean government.

\bibliographystyle{IEEEtran}


\EOD
\end{document}